\documentclass[11pt]{article}

\usepackage[preprint]{acl}

\usepackage{times}
\usepackage{latexsym}
\usepackage[T1]{fontenc}
\usepackage[utf8]{inputenc}
\usepackage{microtype}
\usepackage{inconsolata}
\usepackage{graphicx}

\usepackage{amsmath,amssymb}
\usepackage{algorithm}
\usepackage{algorithmic}
\usepackage{bm}
\usepackage{multirow}
\usepackage{booktabs}
\usepackage[table]{xcolor}
\usepackage{tcolorbox}
\tcbuselibrary{breakable}
\usepackage{bbding}
\usepackage{xurl}
\usepackage{enumitem}
\usepackage{colortbl}
\newcommand*{\inlinegraphics}[1]{%
    \raisebox{-.3\baselineskip}{%
        \includegraphics[
        height=\baselineskip,
        width=\baselineskip,
        keepaspectratio,
        ]{#1}%
    }%
}

\title{
Utilizing and Calibrating Hindsight Process Rewards via Reinforcement \\with Mutual Information Self-Evaluation
}
\author{
    Jiashu Yao\textsuperscript{\rm 1},
    Heyan Huang\textsuperscript{\rm 1},
    Zeming Liu\textsuperscript{\rm 2},
    Yuhang Guo\textsuperscript{\rm 1}\thanks{Corresponding authors.}
    \\
    \textsuperscript{\rm 1}Beijing Institute of Technology \
    \textsuperscript{\rm 2}Beihang University \\
}

\begin{document}
\maketitle
\begin{abstract}
To overcome the sparse reward challenge in reinforcement learning (RL) for agents based on large language models (LLMs), we propose Mutual Information Self-Evaluation (MISE), an RL paradigm that utilizes hindsight generative self-evaluation as dense reward signals while simultaneously calibrating them against the environmental feedbacks. Empirically, MISE enables an agent to learn autonomously from dense internal rewards supplementing sparse extrinsic signals. Theoretically, our work provides the first formal foundation for the paradigm of generative self-rewarding. We prove that utilizing hindsight self-evaluation rewards is equivalent to minimizing an objective that combines mutual information with a KL divergence term between the policy and a proxy reward policy. This theoretical insight then informs and justifies our calibration step, which actively aligns these rewards with the optimal policy. Extensive experiments show that MISE outperforms strong baselines, enabling open-source LLMs about 7B parameters to achieve performance comparable to GPT-4o on validation without expert supervision.
\end{abstract}

\section{Introduction}

Text agent tasks present sequential decision making challenges, employing natural language to describe the environments and the actions \cite{osborne2022survey}. Leveraging large language models (LLMs) \cite{brown2020language,touvron2023llama,achiam2023gpt} agents provides an effective simulation of real-world agent-environment interaction. Recent researches \cite{song-etal-2024-trial,deng2024novice,zeng2025reinforcing} have been exploring fine-tuning LLM agents through reinforcement learning (RL), demonstrating the their potential to learn and adapt in simulated worlds.

\begin{figure}[htb!]
    \centering
    \includegraphics[width=0.9\linewidth]{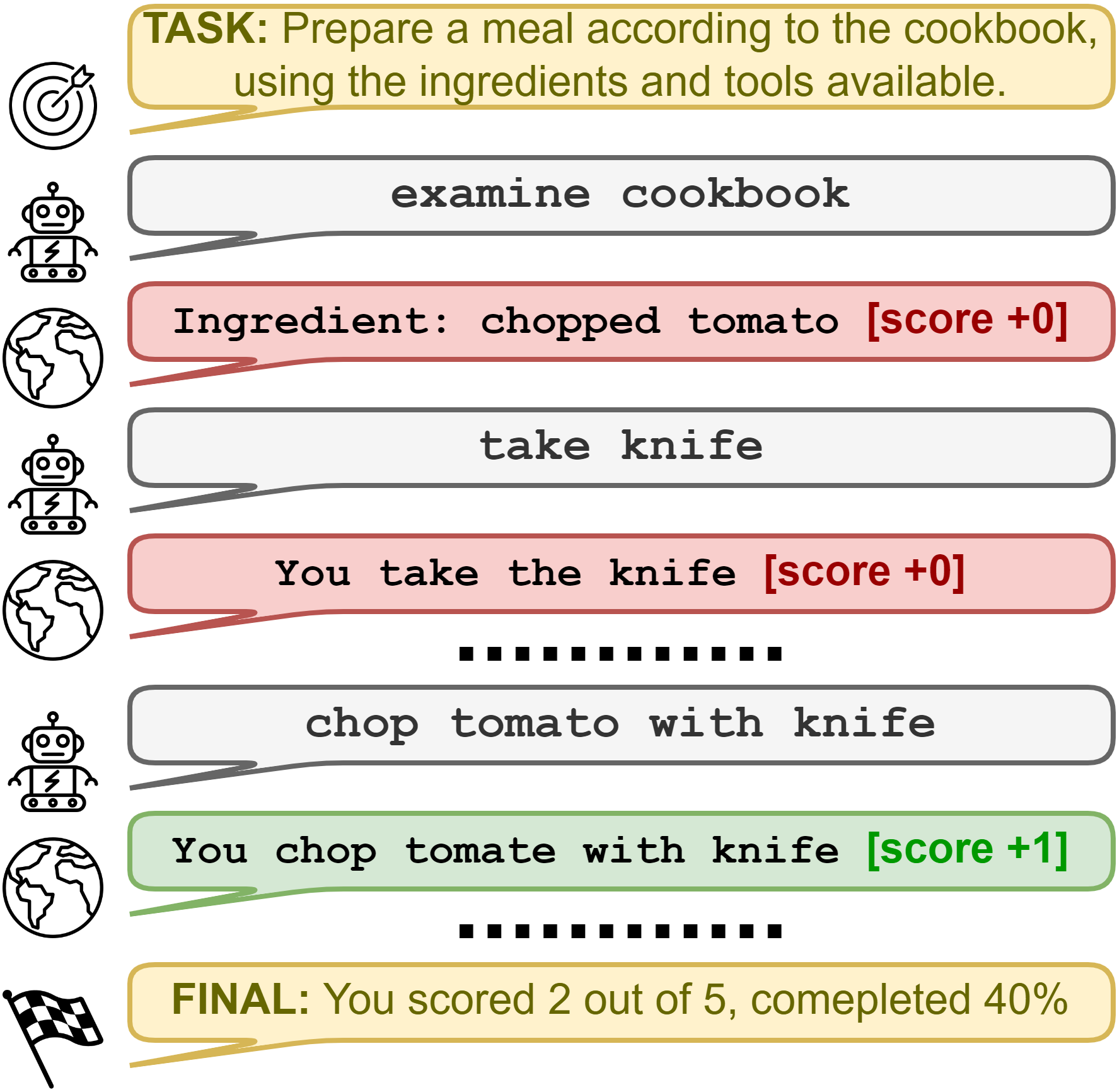}
    \caption{An LLM agent (\inlinegraphics{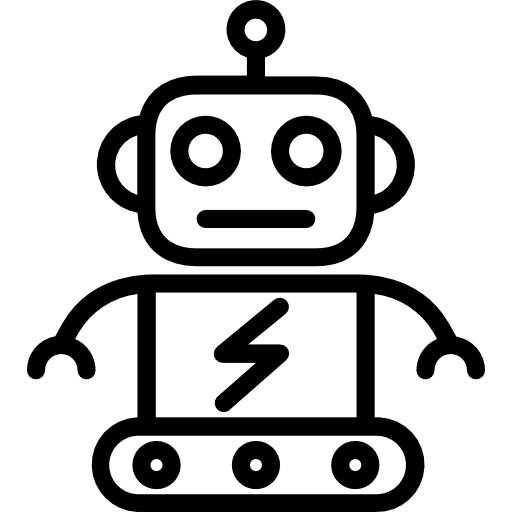}) is tasked with interacting with a environment (\inlinegraphics{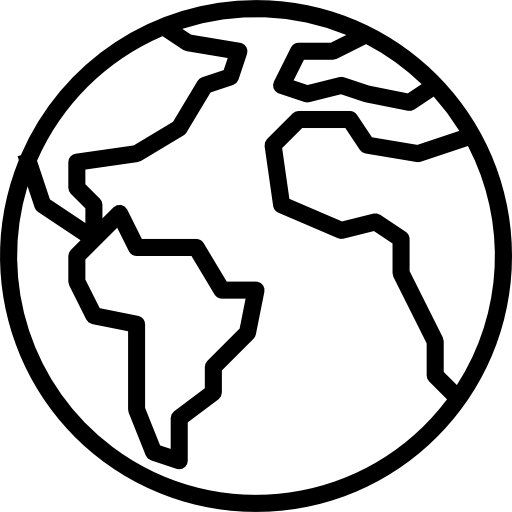}) to complete a task (\inlinegraphics{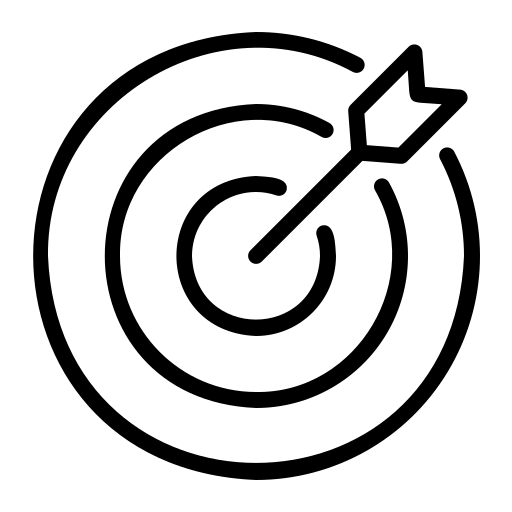}), getting intermediate rewards accumulated to a final score (\inlinegraphics{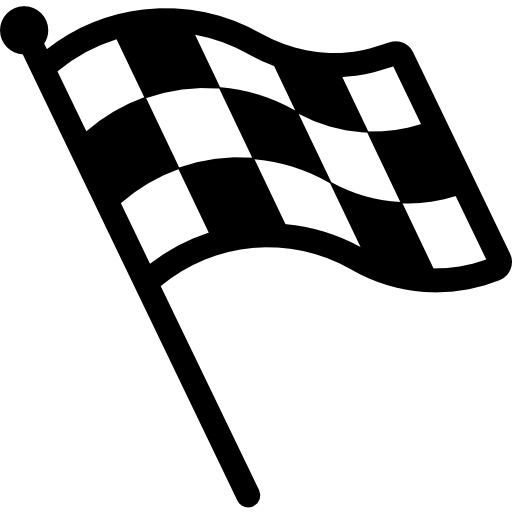}). The environmental rewards are sparse, i.e., most environmental intermediate rewards are zero, so many strategically valuable actions are not properly rewarded.}
    \label{fig:intro}
\end{figure}

One important problem to overcome before achieving agents' autonomously adaptation to unseen tasks through interaction with the environments without supervision \cite{voss2023we,morrisposition}, is the issue of reward sparsity \cite{ladosz2022exploration,deng2024novice}. As exemplified in Figure \ref{fig:intro}, in an agent task, only a few actions directly leading to partial completion of the task (e.g. \texttt{chop tomato with knife}) receive non-zero rewards, while many strategically valuable actions (e.g., \texttt{examine cookbook} and \texttt{take knife}) are not properly rewarded. As a result, in reinforcement learning without step-wise labels, LLMs may suffer from low sample efficiency. Although traditional RL methods have been widely explored
to address reward sparsity, few of them can be easily applied to LLM-based agents, as discussed in Section \ref{sec:related-work}.

\begin{table*}[htbp]
    \centering
    \setlength{\tabcolsep}{1mm}
    \begin{tabular}{lccccc} 
    \toprule
        \textbf{Methods} & \textbf{AI-Feedback} & \textbf{Label-Free} & \textbf{Dense-Reward} & \textbf{Reinforce} & \textbf{Calibrate} \\
        \midrule
        SFT & {\XSolidBrush} & {\XSolidBrush} & {\XSolidBrush} & {\XSolidBrush} & {\XSolidBrush}\\
        PPO \cite{schulman2017proximal} & {\XSolidBrush} & {\CheckmarkBold} & {\XSolidBrush} & {\CheckmarkBold} & {\XSolidBrush}\\
        RFT \cite{yuan2023scaling} & {\XSolidBrush} & {\CheckmarkBold} & {\XSolidBrush} & {\XSolidBrush} & {\XSolidBrush}\\
        DPO \cite{rafailov2024direct} & {\XSolidBrush} & {\XSolidBrush} & {\XSolidBrush} & {\CheckmarkBold} & {\XSolidBrush}\\
        ETO \cite{song-etal-2024-trial} & {\XSolidBrush} & {\XSolidBrush} & {\XSolidBrush} & {\CheckmarkBold} & {\XSolidBrush}\\
        StepAgent \cite{deng2024novice} & {\CheckmarkBold} & {\XSolidBrush} & {\CheckmarkBold} & {\CheckmarkBold} & {\CheckmarkBold} \\
        GRM \cite{mahan2024generative} & {\CheckmarkBold} & {\CheckmarkBold} & {\XSolidBrush} & {\CheckmarkBold} & {\XSolidBrush} \\
        PRM \cite{xi2025agentprm} & {\CheckmarkBold} & {\XSolidBrush} & {\CheckmarkBold} & {\CheckmarkBold} & {\XSolidBrush} \\
        \midrule
        MISE (ours) & {\CheckmarkBold} & {\CheckmarkBold} & {\CheckmarkBold} & {\CheckmarkBold} & {\CheckmarkBold} \\
    \bottomrule
    \end{tabular}
    \caption{Comparison of MISE with the relevant methods. MISE leverages self-evaluation feedback rather than expert supervision to generate informative dense rewards. Meanwhile, MISE calibrates the self-evaluation to keep the dense rewards consistent with environmental rewards, avoiding overfitting to potential evaluation bias.}
    \label{tab:compare}
\end{table*}

To address the challenge of sparse rewards in autonomous adaptation in the era of LLMs, one simple approach is to follow the sucess of generative reward models (GRMs) \cite{ryu2024multi,mahan2024generative,yuan2024self,wu2024meta} and process reward models (PRMs) \cite{choudhury2025process,xi2025agentprm} in reasoning tasks, leveraging the generalized language understanding capabilities inherent in LLMs for step-wise action utility evaluations. Specifically, in agent tasks, an LLM can evaluate the strategic value of its own generated actions at each step, providing a valuable internal reward signals to augment sparse extrinsic rewards.

However, in our concerned LLM agent tasks, such generative process reward models exhibit significant biases if they are not specifically trained with additional data. A specific example detail in Section \ref{subsec:rq3} shows that, current models tend to overvalue the \texttt{inventory} action ($81\%$ cases judged as positive), encouraging the agent to repeatedly check the items currently held rather than making realistic progress. When trained with such biased vanilla PRM, the \texttt{inventory} action constitutes a disproportionately high $24.28\%$ of all generated actions, leading to significant performance drop.

In this work, we fill the gap between the promise of PRMs-assisted LLM agent reinforcement and the significant biases of self-evaluation. We firstly conduct theoretical analysis, and establish the interpretation of hindsight self-evaluation by relating it to minimizing the mutual information (MI) between an agent's action and its future trajectory. The theoretical discussion not only offers deep understanding of the PRMs, but also informs and justifies the RL paradigm design. Empirically, we introduce Mutual Information Self-Evaluation (MISE), a novel reinforcement learning paradigm that integrates sparse environmental rewards with dense internal rewards derived from self-evaluation, while simultaneously applying calibration to avoid potential biases in the self-evaluation process rewards. 

Experiments shows that MISE can significantly improve the performance of LLMs agents, making open-sourced models sizing $\sim$7B exceed GPT-4o-mini and match GPT-4o on valid test. Furthermore, analysis shows that both action selection and self-evaluation components achieve performance improvement, realizing mutual assistance and enhancement within our proposed MISE framework.

Our contributions can be summarized as follows:

\begin{itemize}
    \item We provide a theoretical establishment for hindsight self-evaluation process rewards, demonstrating that using self-evaluation as a reward is related to minimizing the mutual information between the actions and the future trajectory. The theoretical finding then informs and justifies the RL paradigm designs.
    \item For the autonomous adaptation of LLM agents, we propose MISE, an RL paradigm that introduces self-evaluations as dense rewards in addition to sparse environmental rewards, while applying calibration to avoid overfitting to significant evaluation biases.
    \item Extensive experiments and analysis validate the high effectiveness of MISE, and prove the mutual assistance and enhancement effects between agents' action selection and self-evaluation.
\end{itemize}

\section{Related work}
\label{sec:related-work}

\subsection{Adaptation for LLM Agents}

LLM agents are tasked with sequential decision-making problems that use natural languages to convey environment observations and agent actions \cite{osborne2022survey}. These tasks provide a efficient way for agents to learn reasoning, science, morality, etc. \cite{wang2022scienceworld,shi2022stay}. We focus on task-agnostic free-form agents, where no priori components such as admissible command lists or task-specific knowledge graphs \cite{adhikari2020learning,murugesan2021efficient} are involved. In this flexible and challenging scenario, however, current adaptation methods either rely on step-wise supervision or fail to provide dense rewards, as shown in Table \ref{tab:compare}.

Compared to previous methods, our proposed MISE does not rely on additional annotations. Instead, MISE generates dense informative rewards for efficient reinforcement exclusively using the LLM agent alone. Importantly, MISE applies calibration to self-evaluation to keep it consistent with external feedback, avoiding
overfitting to the significant biases.

\subsection{Reward Sparsity}

It has been a major problem in RL that sequential decision making often associates with sparse rewards \cite{ladosz2022exploration}, where only a few actions leading to partial completion of the tasks get non-zero rewards. There has been many successful methods to address reward sparsity before LLM emerges, including curiosity-driven learning \cite{pathak2017curiosity,burda2018large}, auxiliary task optimizing \cite{jaderberg2016reinforcement,riedmiller2018learning}, and hindsight experience replay \cite{andrychowicz2017hindsight}. However, previous methods are not suitable for our concerned tasks, as they either requires priori knowledge or construction specific to the task, or do not fit the LLM RL algorithm (e.g., PPO). Although StepAgent \cite{deng2024novice} proposes to address reward sparsity in LLM agent tasks, it has compulsory requirements for expert annotations.

MISE can also be viewed as an reward sparsity mitigation approach. Its novelty lies in that utilizing and calibrating the superior general language understanding of LLMs for self-evaluation in text agent tasks for the first time, in addition to action selection. This intuitively simple approach is not only empirically effective but theoretically sound.

\subsection{LLM-based Rewarding}

Using LLMs as alternatives to rule-based or expert-based evaluators has been emerging due to their remarkable language capability and cost effectiveness. This trend has led to the emergence of generative reward models (GRMs) \cite{ryu2024multi,mahan2024generative,yuan2024self,wu2024meta}, which is to train a model to align with the internal rewards of another model \cite{lee2024rlaif,sun2024salmon,ahn2024tuning,li2024auto} or itself \cite{ryu2024multi,cao2024enhancing}. However, current LLM evaluators still struggle with issues of unreliability and lack of robustness \cite{gu2024survey}. In the context of LLM agent, there still exists a lack of theoretical grounding and significant vulnerability to evaluation biases \cite{xi2025agentprm,choudhury2025process}.

Other than applying hindsight self-evaluation in LLM agent tasks for the first time, our approach overcomes two shortcoming in existing literature. First, it proposes a robust RL calibration paradigm to prevent overfitting to potential evaluation biases. Second, it mitigates the theoretical unreliability of LLM evaluation by proving its equivalence to minimizing the mutual information term and an inter-policy KL term.

\section{Method}

We firstly formulate hindsight process rewarding in details, then offer a theoretical interpretation of it, and finally propose MISE based on previous discussion.

\subsection{Hindsight Process Reward}

Let $\mathcal{S}$, $\mathcal{A}$, and $\mathcal{O}$ denote state, action and observation space respectively, given a sub-trajectory
\begin{equation}
    \tau_{<t} = <o_0, a_0, ..., o_{t-1}, a_{t-1}, o_t>,
\end{equation}
where $o \in \mathcal{O}$ and $a \in \mathcal{A}$ are previous observations and actions, a policy $\pi_\theta$ sample an action $a_t \in \mathcal{A}$
\begin{equation}
    a_t \sim \pi_\theta(\cdot|\tau_{<t}).
\end{equation}

The environmental reward $r_E(s_t, a_t)$ can evaluate the progress at each step, where $s_t \in \mathcal{S}$ is the state after conducting $\tau_{<t}$. However, $r_E$ is usually very sparse, i.e., most rewards are zero. In these cases, it can be difficult for an agent to efficiently learn from the environment.

To address sparsity in $r_E$, we propose hindsight process reward $r_P$, which assigns a non-zero reward to each action given the whole (including the future) trajectory. Collectively, the optimization target is defined as
\begin{equation}
\label{eq:policy-loss}
    J(\theta)=
    \mathbb{E}_{\tau \sim \pi_\theta} \sum_{t=1}^{T}
    \Big[
    r_E(s_t,a_t) + r_P(\tau_{<t}, a_t,\tau_{>t})
    \Big].
\end{equation}

Note that in Equation \ref{eq:policy-loss}, the process reward $r_P$ is always dependent on the future trajectory $\tau_{>t}$ when evaluating $a_t$.

\subsection{Theoretical Discussion}

Here, we offer a theoretical discussion on the hindsight process reward $r_P$ and show its relation with mutual information. We firstly construct a proxy policy $\pi_P$ to represent the process reward $r_P$ with a Gibbs distribution
\begin{equation}
    \pi_{P}(a_t | \tau_{<t}, \tau_{>t}) = \frac{\pi_{ref}(a_t | \tau_{<t})}{Z(\tau_{<t}, \tau_{>t})} \exp\left( \frac{r_{P}}{\beta} \right),
\end{equation}
where $Z$ is the normalization term, $\pi_{ref}$ is the reference policy, and $\beta$ is the temperature parameter.

We can hereby prove that minimizing the KL divergence between the original and proxy policy is equivalent to maximizing the KL regularized process reward, as detailed in Appendix \ref{app:theory}. Further deriving the KL divergence, we have
\begin{equation}
\begin{aligned}
    &\mathbb{D}_{KL}[\pi_P(a|\tau_{<t}, \tau_{>t}) \parallel \pi(a|\tau_{<t})]\\
    =& \mathbb{E}_{a,\tau_{>t} \sim \pi_P} \Big[ \log \frac{\pi_P(a|\tau_{<t}, \tau_{>t})}{\pi(a|\tau_{<t})} \Big] \\
    =& \mathbb{E}_{a,\tau_{>t} \sim \pi_P} \log \frac{\pi_P(a|\tau_{<t}, \tau_{>t})}{\pi_P(a|\tau_{<t})} + \log \frac{\pi_P(a|\tau_{<t})}{\pi(a|\tau_{<t})} \\
    =& \mathbb{E}_{a,\tau_{>t} \sim \pi_P} \Big[ \log \frac{\pi_P(a, \tau_{>t} | \tau_{<t})}{\pi_P(a|\tau_{<t}) \pi_P(\tau_{>t}|\tau_{<t})} \Big] + \\
    &\mathbb{E}_{a \sim \pi_P} \Big[ \log \frac{\pi_P(a|\tau_{<t})}{\pi(a|\tau_{<t})} \Big]\\
    =& I(a; \tau_{>t} | \tau_{<t}) + \mathbb{D}_{KL}[\pi_P(a|\tau_{<t}) \parallel \pi(a|\tau_{<t})],
\label{eq:mutual-information}
\end{aligned}
\end{equation}
where $I(a; \tau_{>t} | \tau_{<t})$ is the conditional mutual information between the action $a$ and the suffix trajectory $\tau_{>t}$, given the prefix trajectory $\tau_{<t}$.

Equation \ref{eq:mutual-information} shows that, the optimization of hindsight process reward is equivalent to one mutual information term and one KL divergence term. We interpret them respectively as below.

\begin{figure}[htb!]
    \centering
    \includegraphics[width=1.0\linewidth]{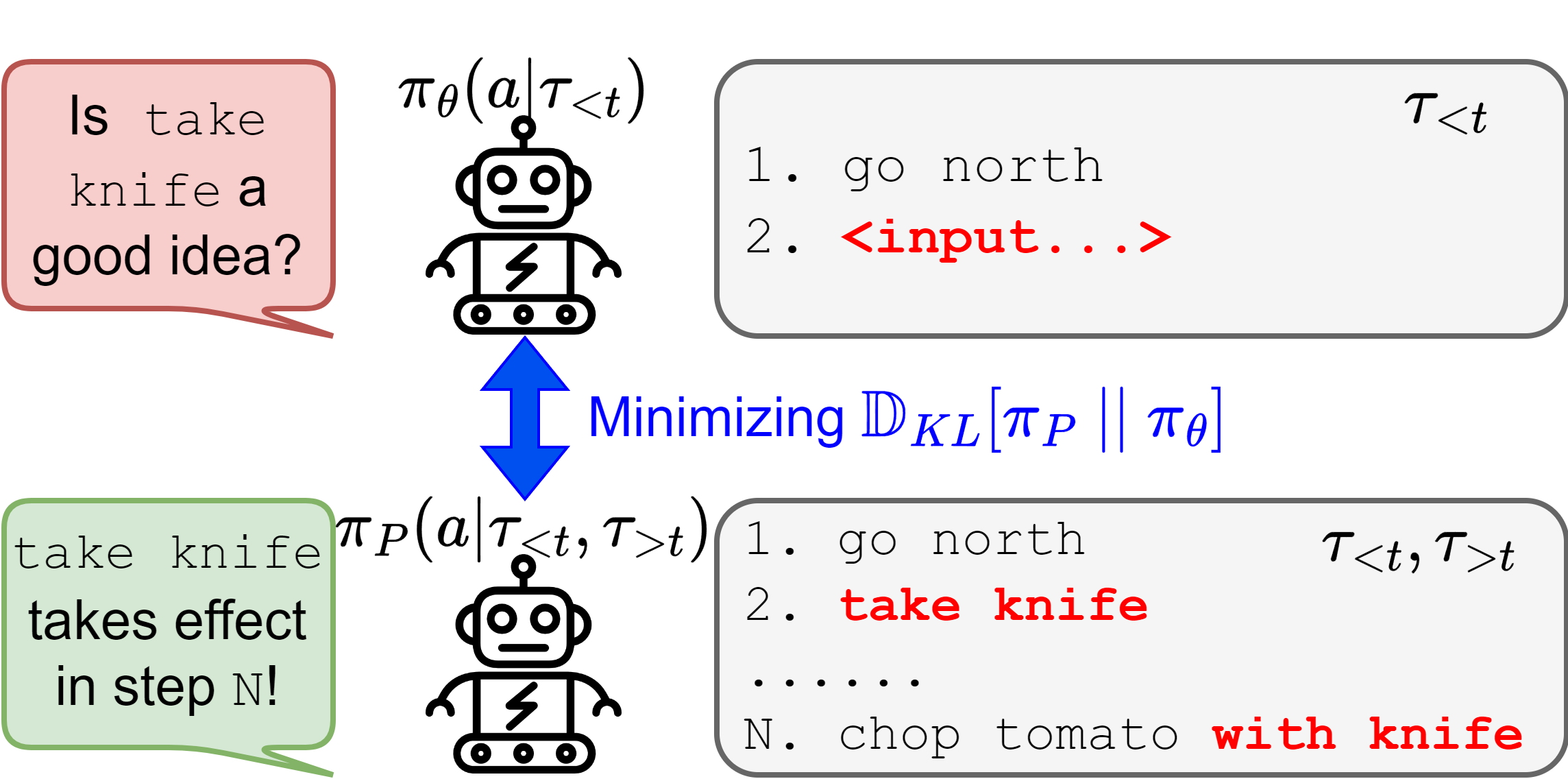}
    \caption{Optimizing towards hindsight self-evaluation ideally enables the agent ($\pi_\theta$ in the top) to make decisions as effective as those made with full future sight ($\pi_P$ in the bottom).}
    \label{fig:mutual-info}
\end{figure}

\paragraph{Mutual information} Minimizing mutual information implies that the agent can select optimal actions in the present without explicitly searching the entire future trajectory.  Ideally, this allows the agent ($\pi_\theta$) to make choices that are as effective as those made with full knowledge of the future ($\pi_P$). As is shown in the example of Figure \ref{fig:mutual-info}, an LLM-based policy $\pi_\theta$ may not be able to choose the right action in the first place. However, optimizing toward $r_P$ pulls its distribution closer to the proxy policy $\pi_P$, which can see the entire trajectory and find the utility of the action \texttt{take knife}.

\paragraph{KL divergence} Different from the mutual information term, the KL divergence term is to let policy $\pi_\theta$ mimic a marginalized distribution $\pi_P(a|\tau_{<t})$, without introducing effective future information such as $\tau_{>t}$. Since $\pi_P(a|\tau_{<t})$ is not explicitly calibrated during vanilla RL training, the KL divergence term may lead to biases in the optimization target. The KL divergence term informs the need of actively calibrating self-evaluations for a reliable process rewarding.

\subsection{Mutual Information Self-Evaluation (MISE)}

\begin{figure}[htb!]
    \centering
    \includegraphics[width=0.95\linewidth]{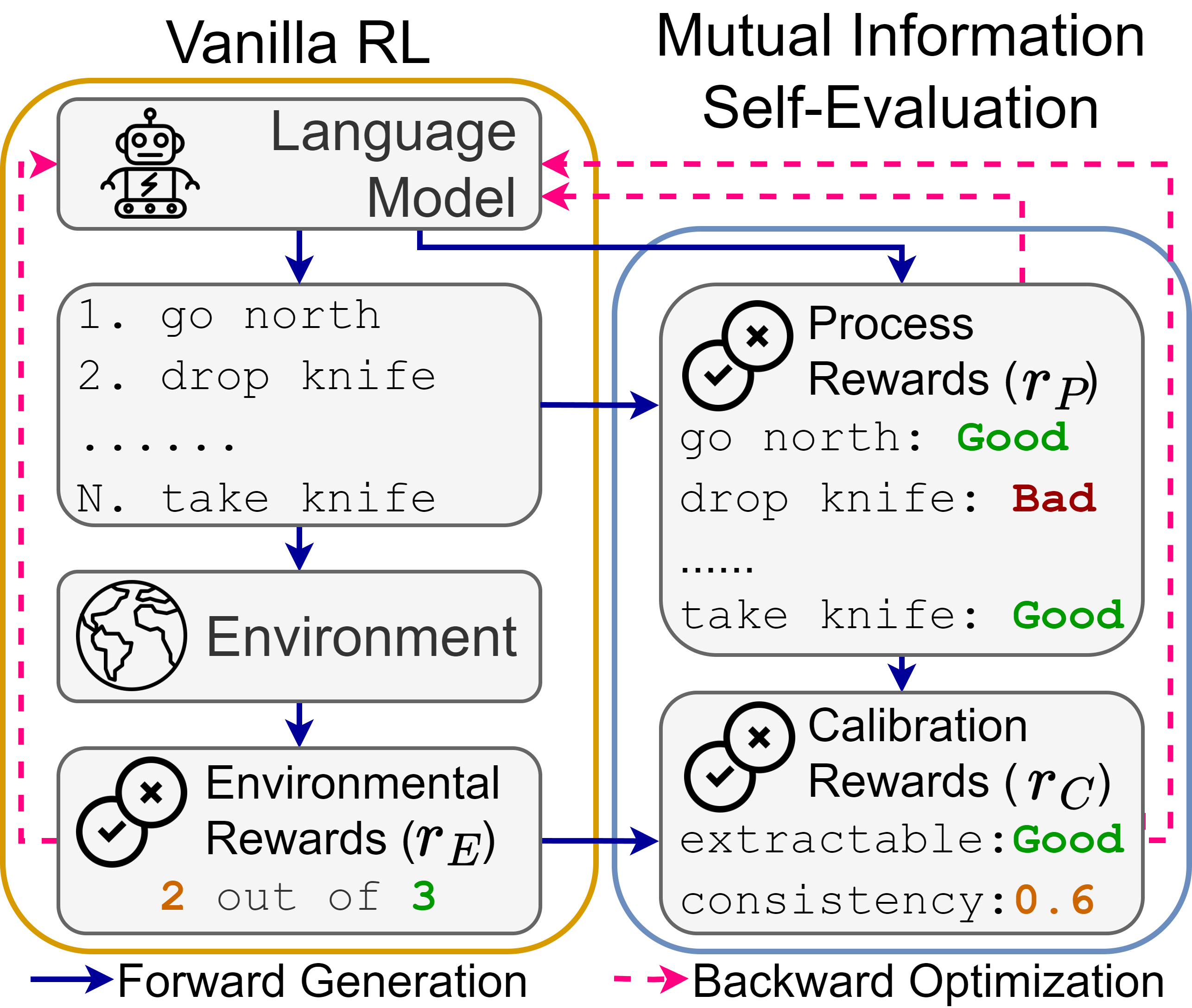}
    \caption{MISE framework overview. Apart from environmental sparse rewards ($r_E$), MISE generates dense process self-evaluation rewards ($r_P$), which are calibrated by rewards ($r_C$) to be consistent with actual performance. MISE adapts LLM agents with dense rewards without overfitting to self-evaluation biases.}
    \label{fig:method}
\end{figure}

Based on the discussion above, we propose mutual information self-evaluation (MISE), an RL paradigm that utilizes the mutual information to address sparsity, and calibrates the process rewards to avoid self-evaluation biases. As is shown in Figure \ref{fig:intro}, in additional to the sparse environmental reward $r_E$, MISE adds two rewards to the vanilla RL process, namely, the reward of hindsight process self-evaluation $r_P: \tau_{<t} \times a \times \tau_{>t} \rightarrow \{-1, 1\}$ and the calibration reward $r_C: \tau \times r_c \rightarrow [-1, 1]$.

\paragraph{Hindsight process rewards} $r_P$ directly addresses the reward sparsity problem, as it uses the general language understanding ability of LLMs to assign a non-zero score to each action given the whole trajectory. Specifically, the description of the whole trajectory is given in the prompt as shown in Appendix \ref{app:prompts}, and an LLM is instructed to evaluate the strategic value of each action within the context of the entire trajectory to assign either $-1$ or $1$.

\paragraph{Calibration rewards} $r_C$, on the other hand, regularizes the consistency between the self-evaluation and the real performance, to prevent LLMs from overfitting the potentially biased rewards. We design heuristic scores as $r_C$,
\begin{equation}
    r_C = \text{clip}(1.2 - 2 \times |p_{\text{real}}-p_{\text{self-eval}}| ,0, 1) \times 2 - 1,
\label{eq:cabiliration-reward}
\end{equation}
where given a trajectory $\tau$ and the maximum environmental score $S$, $p_{\text{real}} \in [0, 1]$ and $p_{\text{self-eval}} \in [0, 1]$ are defined as
\begin{equation}
\begin{aligned}
    p_{\text{real}} &= \frac{\sum_{t=1}^{T} r_E(s_t,a_t)}{S},\\
    p_{\text{self-eval}} &= \frac{\sum_{t=1}^{T} r_P(\tau_{<t},a_t,\tau_{>t})}{T}.
\end{aligned}
\end{equation}

Generally, $r_C$ is aimed to make the self-evaluation of LLMs on a trajectory consistent to its real performance. A detailed discussion and justification of the $r_C$ design is in Appendix \ref{app:rm}.

\paragraph{Algorithm}

Combining Equation \ref{eq:policy-loss} with the two proposed rewards $r_P$ and $r_C$, the optimization target of MISE is 
\begin{equation}
\label{eq:our-optimization}
    J(\theta)=
    \mathop{\mathbb{E}}_{\tau \sim \pi_\theta}
    \Big[
    \sum_{t=1}^{T} \frac{r_E + r_P}{2}
    + r_C
    \Big]
    - \beta \mathbb{D}_{KL}.
\end{equation}



\section{Experiments}

\begin{table*}[htb!]
    \setlength{\tabcolsep}{1.5mm}
    \centering
    \begin{tabular}{lccc|ccccccc}
        \toprule
        \multirow{2}{*}{\textbf{Models}} & \multicolumn{3}{c|}{\textbf{Average}} & \multicolumn{2}{c}{\textbf{FTWP}} & \multicolumn{2}{c}{\textbf{ScienceWorld}} & \multicolumn{2}{c}{\textbf{WebShop}} \\
        \cmidrule(lr){2-4} \cmidrule(lr){5-6} \cmidrule(lr){7-8} \cmidrule(lr){9-10} & FF Rate & Suc (V) & Suc (T) & Suc (V) & Suc (T) & Suc (V) & Suc (T) & Suc (V) & Suc (T) \\
        \midrule

        GPT-4o          & 82.13 & 40.40 & 38.29 & 43.66 & 51.08 & 40.65 & 32.41 & 36.88 & 31.39 \\
        GPT-4o-mini     & 72.54 & 32.55 & 29.64 & 27.26 & 31.63 & 38.60 & 31.98 & 31.79 & 25.30 \\
        \midrule

        LLaMA3-8B       & 67.83 & 28.92 & 23.06 & 21.73 & 17.24 & 36.12 & 31.79 & 28.90 & 20.14 \\
        \  + ReAct      & 68.68 & 28.30 & 23.63 & 18.81 & 17.80 & 37.97 & 33.10 & 28.13 & 19.98 \\
        \  + RFT        & 77.47 & 31.29 & 26.10 & 24.75 & 22.32 & 38.64 & 34.02 & 30.48 & 21.97 \\
        \  + online DPO & 67.47 & 29.54 & 24.40 & 22.54 & 20.09 & 36.76 & 32.14 & 29.31 & 20.96 \\
        \  + PPO        & 78.90 & 28.20 & 22.33 & 27.87 & 24.48 & 26.04 & 20.03 & 30.69 & 22.47\\
        \  + PRM        & \textbf{80.70} & 32.75 & 27.36 & 28.45 & 25.61 & 38.87 & 33.65 & 30.94 & 22.81\\
        \  + MISE       & 77.57 & \textbf{36.48} & \textbf{30.47} & \textbf{34.71} & \textbf{28.90} & \textbf{43.40} & \textbf{38.18} & \textbf{31.34} & \textbf{24.34} \\
        \midrule

        Qwen2-7B  & 44.85 & 19.60 & 15.72 & 16.20 & 13.11 & 12.35 & 10.67 & 30.26 & 23.38\\
        \  + PPO  & 71.34 & 27.86 & 22.15 & 30.69 & 20.85 & 22.09 & 21.85 & 30.80 & 23.75\\
        \  + PRM  & 70.92 & 28.28 & 22.95 & 31.80 & 22.81 & 22.60 & 22.24 & 30.43 & 23.81\\
        \  + MISE & \textbf{73.14} & \textbf{32.12} & \textbf{26.22} & \textbf{33.80} & \textbf{24.90} & \textbf{28.87} & \textbf{28.61} & \textbf{33.68} & \textbf{25.15}\\
        \midrule

        Gemma2-9B & 61.23 & 9.93 & 9.74 & 16.30 & 17.30 & 13.49 & 11.91 & 0.0 & 0.0\\
        \  + PPO  & 86.42 & 25.87 & 20.67 & 36.95 & 29.83 & 40.66 & 32.19 & 0.0 & 0.0\\
        \  + PRM  & 88.90 & 26.63 & 21.45 & 37.77 & 30.50 & 42.12 & 33.85 & 0.0 & 0.0\\
        \  + MISE & \textbf{89.74} & \textbf{28.07} & \textbf{23.25} & \textbf{40.14} & \textbf{32.94} & \textbf{44.08} & \textbf{36.80} & 0.0 & 0.0\\

        \bottomrule
    \end{tabular}
    \caption{Main experiments on three tasks. Bolded numbers denote the best results for each base model.}
    \label{tab:main-exp}
\end{table*}

\subsection{Test bench}

\paragraph{Task}

Our main experiments and analyses are conducted on First TextWorld Problems (FTWP) \cite{FirstTextWorldProblems2019}. This task is built on TextWorld \cite{cote18textworld}, with the aim of building an autonomous agent that can navigate and interact within a text-simulated modern house and finally cook a meal. For generalizability, we conduct experiments on other agent tasks including ScienceWorld \cite{wang2022scienceworld} and WebShop \cite{yao2022webshop}.

\begin{figure}[htbp]
    \centering
    \includegraphics[width=0.8\linewidth]{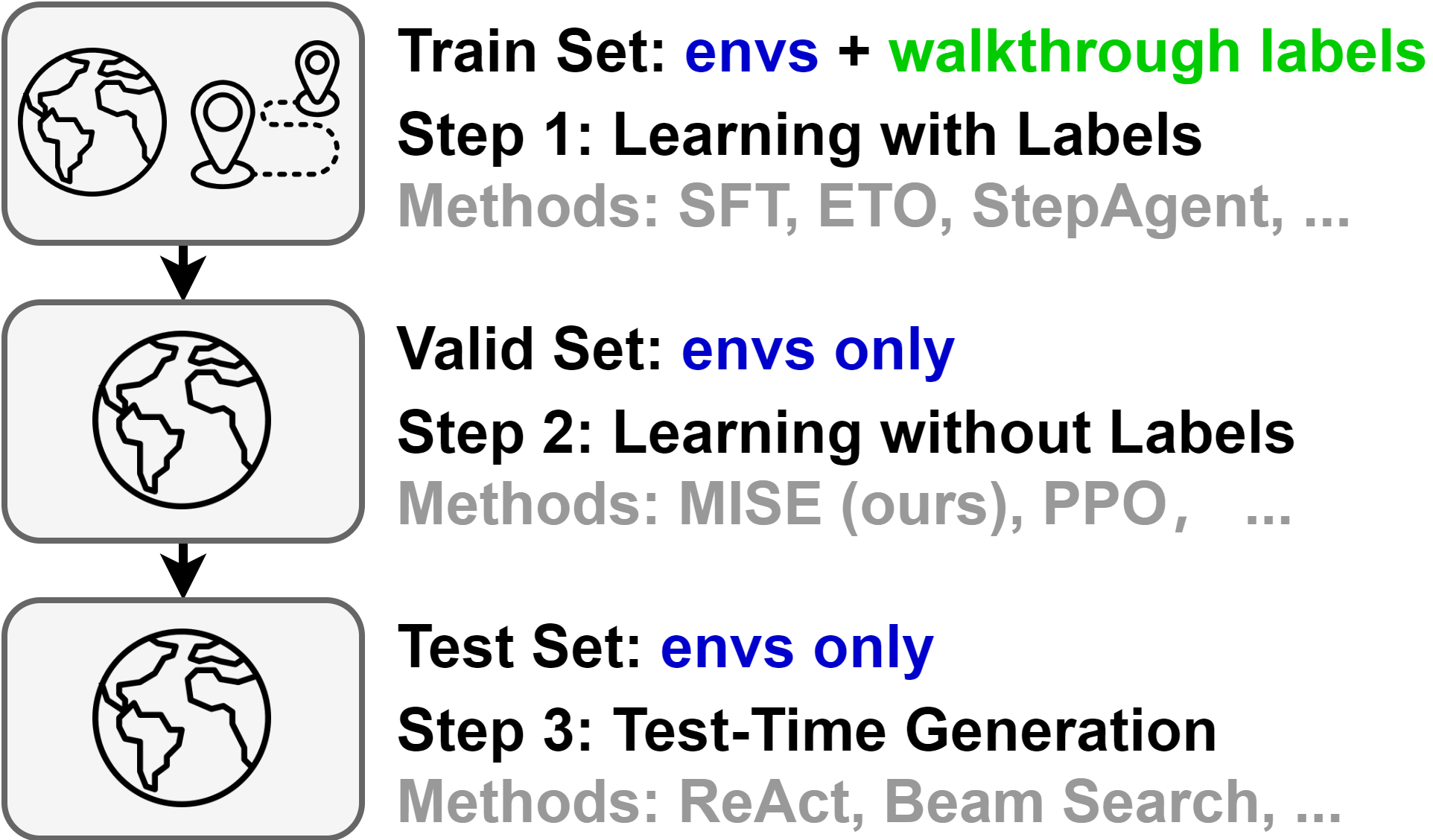}
    \caption{The overall experimental procedure. Each individual steps can be included or skipped.}
    \label{fig:testbench}
\end{figure}

\paragraph{Procedure}

The experimental procedure is illustrated in Figure \ref{fig:testbench}. The process consists of three stages. First, and optionally, LLMs can learn through imitation of expert demonstrations. SFT methods and those requiring data annotations fall within this initial stage. Second, LLMs learn solely through interaction with the environment in a completely label-free manner. This second stage embodies the concept of autonomy of adaptability and is the context in which our proposed MISE operates. Third, LLMs are evaluated on unseen environments to assess the generalizability.

\paragraph{External rewards}

The external rewards refer to the scores provided by environments indicating the completion of a sub-goal, which are sparse as most are zeros, neglecting many strategically valuable actions. The step $j$-th reward for action $r_{Ej}$ is formulated as the average of the suffix scores:
\begin{equation}
\label{eq:ir}
\begin{aligned}
     r_{Ej} &= -1+2 \times [ \sum_{i=j}^{T} s_i \ / \ (T-j+1)],
\end{aligned}
\end{equation}
where $s_i$ denotes the environmental score of step $i$.

\paragraph{Metrics}

The performance of LLMs are evaluated with micro success rate (Suc Rate). Suc Rate indicates the degree of task completion, i.e., the rate at which the agent achieves the given goal. We also report the format faithfulness rate (FF Rate) \cite{yao2024reff}, which refers to the percentage of commands that can be compiled and executed by the environment, representing the degree to which LLMs can obey the given command format.

\subsection{Experimental setup}

The key experimental setup are listed here, and more details are described in Appendix \ref{app:input-representation} and \ref{app:implementation}.

\paragraph{Models}

We conduct experiments on LLaMA3-8B-Instruct, Qwen2-7B-Instruct, and Gemma2-9B-Instruct. We mainly conduct comparison experiments on LLaMA3, as its original model performs best. We also compare MISE with close-sourced models including GPT-4o and GPT-4o-mini.

\paragraph{Baselines}

We compare our autonomous adaptation method MISE to other strong label-free baselines. (1) ReAct \cite{yaoreact} encourages LLMs to think before act using prompting techniques. (2) RFT \cite{yuan2023scaling} collects the successful sampled trajectories to construct a dataset used for SFT LLMs. (3) Online DPO \cite{rafailov2024direct} optimizes LLMs by training on pairwise comparisons of preference data, which is generated by the models online. (4) PPO \cite{schulman2017proximal}, the de-facto standard for RL in LLMs, directly optimizes the LLMs for higher environmental rewards. (5) PRM \cite{mahan2024generative} uses LLM self-evaluation as additional rewards based on vanilla PPO.

We further experiment whether annotation-free MISE can take effects in scenarios with trajectory labels by post-tuning adapted models, comparing with methods requiring labels, including: (1) SFT, (2) DPO-based methods like ETO \cite{song-etal-2024-trial} and StepAgent-Implicit \cite{deng2024novice}.

\paragraph{Hyper-parameters}

We employ commonly used hyperparameters without cherry-picking to ensure the reliability of the experiments. In all LLM adaptations, we use LoRA \cite{hu2022lora} with rank being $64$. SFT uses a linear scheduled learning rate $3e-6$, trains for $1$ epoch with batch size $256$. DPO uses a linear scheduled learning rate $3e-7$, trains for $1$ epoch with batch size $32$. PPO uses a constant learning rate $1e-5$, trains for $3$ epochs with batch size being $32$, KL coefficient ($\beta$ in Equation \ref{eq:our-optimization}) being $0.1$. All inferences use greedy decoding.

\subsection{Results}

The results in the label-free scenario on FTWP, ScienceWorld, and WebShop tasks are shown in Table \ref{tab:main-exp}. Without any external labels, MISE significantly improves the task completion rate, outperforming previous methods like RFT, online DPO, PPO, and PRM by a large margin, matching or even exceeding GPT-4o and GPT-4o-mini. Notably, the superiority of MISE over PRM validates the effectiveness of calibrating self-evaluations with the calibration rewards ($r_C$).

In the experimental results, we have two outlier observations, including (1) PPO performance degradation with LLaMA3 on ScienceWorld, and (2) less improvement of all methods on WebShop compared to on ScienceWorld. The first observation aligns with findings from previous work \cite{song-etal-2024-trial}. For the second observation, we speculate this is because the average trajectory length in WebShop ($\sim5$) is significantly shorter than in the other two tasks ($\sim19$), making process optimization more challenging. As a notable specific case, both the original and RL-finetuned Gemma models fail to follow the online shopping format instructions, resulting in no success on WebShop.

\begin{table}[htb!]
    \setlength{\tabcolsep}{1mm}
    \centering
    \begin{tabular}{lccc}
        \toprule
        \textbf{Models} & FF Rate & Suc(Val) & Suc (Test) \\

        \midrule
        GPT-4o          & 82.13 & 43.66 & 51.08  \\
        GPT-4o-mini     & 72.54 & 27.26 & 31.63  \\

        \midrule
        SFT & \textbf{86.46} & 66.54 & 58.53 \\
        ETO  & 84.80 & 67.04 & 59.81 \\
        StepAgent & 84.66 & 67.21 & 59.69 \\
        MISE & 84.79 & \textbf{68.45} & \textbf{62.66} \\
        
        \bottomrule
    \end{tabular}
    \caption{Experiment results on FTWP task in the scenario with training labels. All adaptations are based on LLaMA3-8B. Additional details for SFT are listed in Appendix \ref{app:implementation}.}
    \label{tab:supervised-exp}
\end{table}

Although MISE is an label-free RL method aiming for autonomy of adaptability, it can also improves the performance of SFT-tuned models through post-tuning. As shown in Table \ref{tab:supervised-exp}, MISE based on a SFT model outperforms both its reference model and other RL-based methods, and surpassing strong close-sourced models such as GPT-4o. A discussion on how MISE can help SFT models improve is shown in Appendix \ref{app:insights}.

\begin{table*}[htbp]
    \centering
    \begin{tabular}{lccccc}
        \toprule
        \multirow{2}{*}{\textbf{Models}} & \multicolumn{2}{c}{\textbf{Self-Evaluation Quality}} & \multicolumn{3}{c}{\textbf{Task Completion Quality}} \\
        \cmidrule(lr){2-3} \cmidrule(lr){4-6} & Extractability & Accuracy & FF Rate & Suc Rate (Val) & Suc Rate (Test) \\
        \midrule
        GPT-4o                        & 97.73 & 85.22 & 82.13 & 43.66 & 51.08 \\
        \midrule
        LLaMA3                        & 79.55 & 67.28 & 67.83 & 21.73 & 17.24 \\
        LLaMA3 + PPO                  & 75.00 & 68.13 & 78.90 & 27.87 & 24.48 \\
        LLaMA3 + MISE                 & \textbf{86.36} & \textbf{72.33} & 77.57 & \textbf{34.71} & \textbf{28.90} \\
        \ \ + w/o $r_m$ (vanilla GRM) & 68.18 & 65.91 & \textbf{80.86} & 28.45 & 25.61 \\
        \ \ + random eval             & 65.45 & 64.20 & 77.84 & 23.69 & 19.03 \\
        \ \ + flipped eval            & 2.49 & 57.89 & 2.81 & 0.47 & 0.24 \\
        
        \bottomrule
    \end{tabular}
    \caption{Analytic experiments with optional corruption of self-evaluation process. Results of both self-evaluation and task completion are assessed. Note that MISE without the reward for self-evaluation $r_C$ is equivalent to the vanilla PRM approach. Best results based on LLaMA3 models are shown in bold.}
    \label{tab:analysis-exp}
\end{table*}

\section{Analasis}

Since our proposed MISE method centers on leveraging self-evaluation to improve autonomy of adaptability, we are particularly interested in a deeper understanding of this evaluation process. In this section, we present three analyses to examine the mutual assistance and enhancement effects in LLM agents between the action selection and the self-evaluation components: (1) Are step-wise evaluations generated by LLMs of high quality? (2) Does the self-evaluation help improve LLMs' action selection? (3) How does the calibration for process rewarding ($r_C$) regularization affects agent action selection?

\subsection{Quality of self-evaluation}
\label{subsec:rq1}

MISE is to use self-evaluation as dense internal rewards to improve action selection, while simultaneously calibrate the self-evaluation process by optimizing the consistency between self-evaluation and the actual task completion rate. Ideally, after training with MISE, LLMs will excel at both action generation and self-evaluation. We firstly analyze the quality of the LLMs' self-evaluations.

In doing so, we first construct a evaluation dataset. we randomly select $44$ instances which contains $764$ actions, and manually label each action to determine whether it is strategically useful for the ultimate goal. Then we test LLMs on the datasets with two metrics, (1) extractability, i.e., whether the generated evaluation can pull out a meaningful score, and (2) accuracy, i.e., whether the extracted evaluation is the same as human-given labels.

The results are shown in Table \ref{tab:analysis-exp}. It is evident that with the rewards for self-evaluation $r_C$, MISE can improve the evaluation capability in the context of LLM agents. Meanwhile, without those rewards punishing self-evaluations for being inconsistent with actual results, it can suffer from a decline due to catastrophic forgetting \cite{luo2023empirical}.

Moreover, from the perspective of probably approximately correct (PAC) learnability \cite{valiant1984theory,angluin1988learning}, the existence of noise in evaluations or labels necessitates larger training datasets. Given that the self-evaluations produced by LLaMA3-based agents are not as accurate as those from larger models such as GPT-4o, there is significant room for improvement. While the performance margin between MISE and previous methods is already substantial, it has a great potential to obtain even better performance when employing more powerful base models.

\subsection{Impact of self-evaluation on acting}
\label{subsec:rq2}

Having examined the quality of self-evaluation, we now turn to another crucial question: to what extent does the self-evaluation quality influences LLM action selection efficacy?

To investigate this, we progressively corrupt the self-evaluations at three increasing levels of severity. (1) Training without rewards for self-evaluation: as demonstrated in Section \ref{subsec:rq1}, this scenario results in a slight degradation of evaluation accuracy. (2) Training with random self-evaluations: this scenario introduces noise into the training process. (3) Training with flipped self-evaluations: this scenario represents a more severe corruption, where the evaluator deliberately provides false evaluations.

The results are presented in Table \ref{tab:analysis-exp}. With the first level of corruption (training without rewards for self-evaluation), the model experiences a performance decrease, yet still outperform PPO, as even the corrupted self-evaluations retain some useful dense information. At the second level (training with random self-evaluations), the model's performance falls below that of PPO due to the introduced random noise, but still surpasses the original LLaMA3 as the environment continues to provide some sparse feedback. At the third level of corruption (flipped self-evaluations), the model is unable to learn effectively, instead generating nonsensical outputs that attempt to conform to the deliberately misleading evaluations.

\subsection{Rewards for self-evaluation}
\label{subsec:rq3}

MISE is composed of three rewards, namely environmental rewards ($r_E$), self-evaluation process rewards ($r_P$), and self-evaluation calibration rewards ($r_C$). Among these, $r_C$ is unique in that it does not directly optimize the action selection process. Instead, it focuses on calibrating the self-evaluation. Nevertheless, as demonstrated in the preceding two subsections, $r_C$ exhibits a significant positive influence on both self-evaluation quality and action selection performance.

To better understand the crucial role of $r_C$ in the MISE training process, we examine the training-time task completion rate and the positive self-evaluations rate epoch by epoch, comparing models trained with $r_C$ (our proposed MISE) to those without it (vanilla PRM).

\begin{figure}[htbp]
    \centering
    \includegraphics[width=1.0\linewidth]{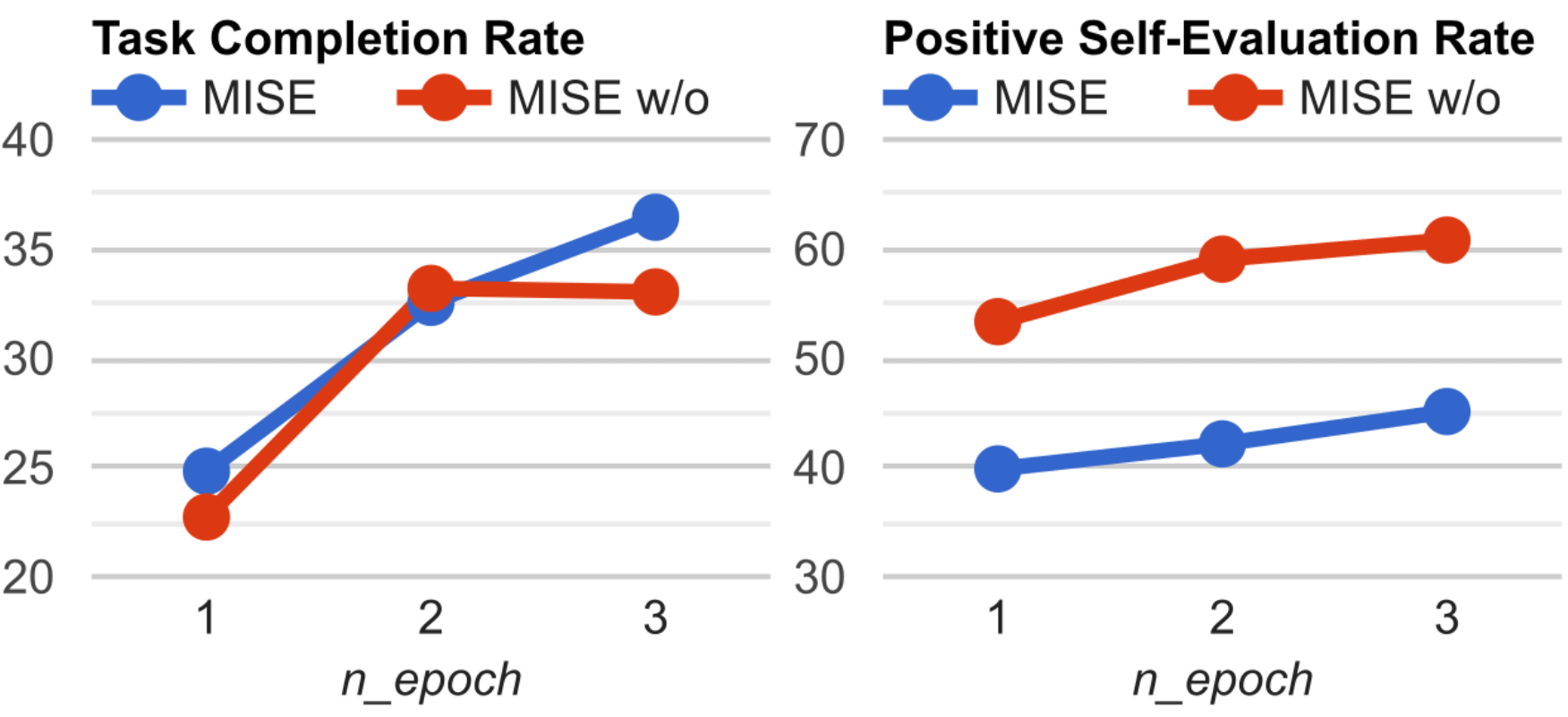}
    \caption{The task completion rates and positive self-evaluation rates in three epochs on the valid set. MISE w/o denotes MISE without self-evaluation calibration rewards.}
    \label{fig:eval-bias}
\end{figure}

The results are illustrated in Figure \ref{fig:eval-bias}. Without $r_C$, the LLMs exhibit overconfidence in their action selections. Although, during training, MISE without $r_C$ lags behind MISE with $r_C$ in terms of actual task completion rates, it assigns itself significantly higher self-evaluations (up to 60\% positive).

The underlying reason for the overconfidence is the inherent significant bias of LLM self-evaluations. Self-evaluation is not a perfect reflection of the real action plausibility within given environments. Instead, it can favor certain actions that are not actually useful for the completion of the task. While self-evaluation effectively addresses reward sparsity, it as a part of optimization target can mislead the agent to generate actions that are perceived as good by itself but detrimental to performance. On the other hand, the self-evaluation calibration rewards ($r_C$) can help avoid the overfitting to potential biases by rcalibrating the self-evaluation to be consistent with the real completion rate. As a result, MISE can simultaneously mitigate reward sparsity and self-evaluation biases.

A concrete example of this bias is the tendency of LLaMA3-8B to overvalue the \texttt{inventory} action, which checks the items currently held by the agent.  Without $r_m$, the \texttt{inventory} action constitutes a disproportionately high $24.28\%$ of all generated actions. However, with the inclusion of $r_m$, this percentage decreases to $14.34\%$, only slightly higher than the $8.51\%$ in the original model.

\section{Conclusion}

In this work, we present MISE, an RL paradigm for LLM autonomous agent adaptations. MISE encourages the agent to learn from its self-evaluation but not conform to the potential bias, by simultaneously utilizing and calibrating generative self-evaluation rewards, which is theoretically justified by its relation to minimizing the mutual information. Extensive experiments validate the high effectiveness of MISE with or without step-wise labels across diverse models and tasks. Moreover, analyses prove the mutual assistance and enhancement effects in LLM agents between the action selection and the self-evaluation components, suggesting the potential of self-evaluation in autonomous adaptation. MISE facilitates efficient and effective autonomous agent adaptation without intensive expert labeling, presenting a step towards autonomy of adaptability in the reinforcement of LLM agents.

\bibliography{custom}

\clearpage

\appendix

\section{Construction of Proxy Policy}
\label{app:theory}

\begin{figure*}[htb!]
    \centering
    \includegraphics[width=1.0\linewidth]{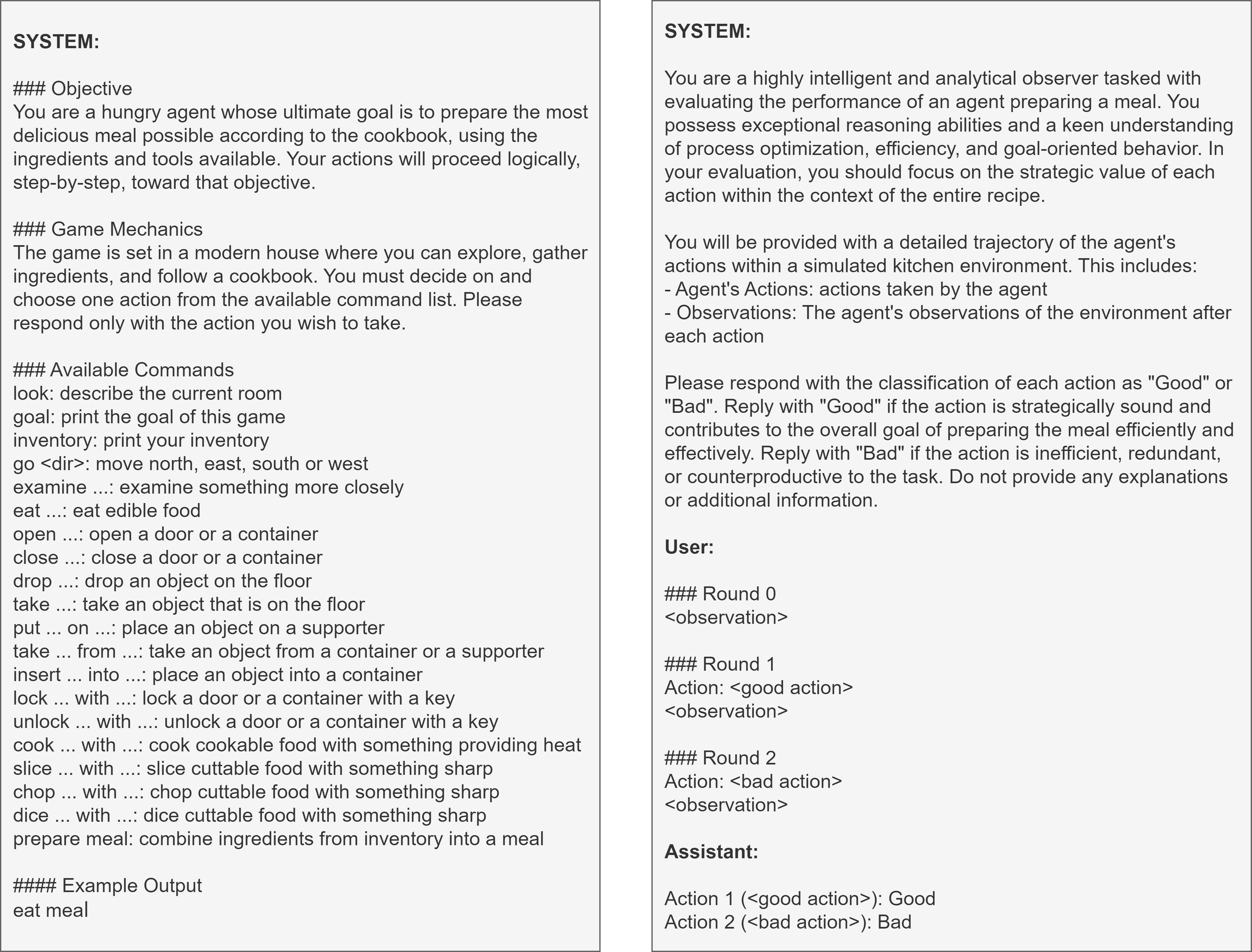}
    \caption{Prompts for agents (left) and self-evaluation (right).}
    \label{fig:prompts}
\end{figure*}

Given the definition of $\pi_P$
\begin{equation}
\begin{aligned}
    &\pi_{P}(a_t | \tau_{<t}, \tau_{>t}) =\\
    & \frac{\pi_{ref}(a_t | \tau_{<t})}{Z(\tau_{<t}, \tau_{>t})} \exp\left( \frac{r_{P}(\tau_{<t}, a_t, \tau_{>t})}{\beta} \right),
\end{aligned}
\end{equation}
take the logarithm of both sides of the equation,
\begin{equation}
\begin{aligned}
    &\log \pi_P(a_t | \tau_{<t}, \tau_{>t}) = \\
    &\log \pi_{ref}(a_t | \tau_{<t}) + \frac{r_P(\tau_{<t}, a_t, \tau_{>t})}{\beta}\\
    &- \log Z(\tau_{<t}, \tau_{>t}),
\end{aligned}
\end{equation}
and substitute it into the KL expansion, we have
\begin{equation}
\begin{aligned}
    &\mathbb{D}_{KL}(\pi \parallel \pi_P)\\
    &= \mathbb{E}_{\tau \sim \pi} [ \log \pi(a_t | \tau_{<t}) - \log \pi_P(a_t | \tau_{<t}, \tau_{>t}) ]\\
    &= \mathbb{E}_{\tau \sim \pi} \Big[ \log \frac{\pi(a_t | \tau_{<t})}{\pi_{ref}(a_t | \tau_{<t})}\Big]\\
    &\quad - \frac{1}{\beta} \mathbb{E}_{\tau \sim \pi} r_P(\tau_{<t}, a_t, \tau_{>t})
     + \log Z(\tau_{<t}, \tau_{>t})\\
    &= \mathbb{D}_{KL}(\pi \parallel \pi_{ref}) - \frac{1}{\beta} \mathbb{E}_{\tau \sim \pi} r_P(\tau_{<t}, a_t, \tau_{>t})\\
    &\quad + \log Z.
\end{aligned}
\end{equation}

Since $Z$ is not related to the policy to optimize ($\pi$), we have
\begin{equation}
\begin{aligned}
    &\min_{\pi} \mathbb{D}_{KL}(\pi \parallel \pi_P)
    \iff\\
    &\max_{\pi}
    \Big[ \mathbb{E}_{\tau \sim \pi} r_{P} - \beta \mathbb{D}_{KL}(\pi \parallel \pi_{ref}) \Big].
\end{aligned}
\end{equation}

\section{Prompts}
\label{app:prompts}

The prompts used are shown in Figure \ref{fig:prompts}.

\section{Discussion on Heuristic Reward Design}
\label{app:rm}

\begin{figure*}[htb!]
    \centering
    \includegraphics[width=1.0\linewidth]{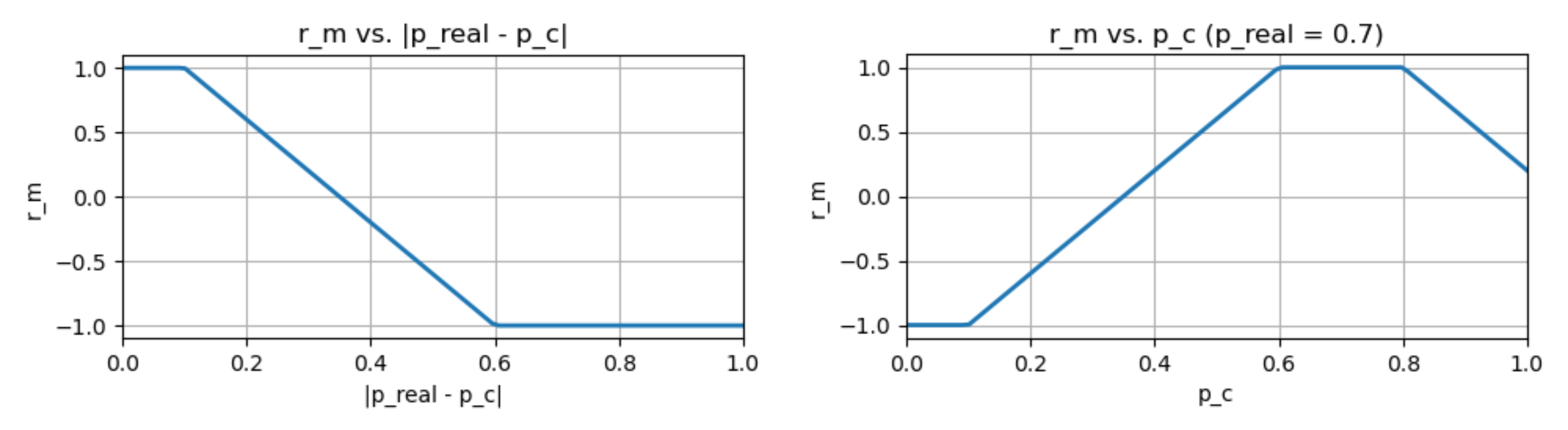}
    \caption{The graph (left) and illustration (right) for reward for self-evaluation ($r_m$). $r_m$ punishes $p_\text{real}$ and $p_c$ for being inconsistent, while not requiring them to be perfectly identical.}
    \label{fig:heuristic}
\end{figure*}

The self-evaluation calibration reward ($r_C$) is designed to prevent self-evaluation from deviating significantly from actual environmental feedback, thereby preventing the LLM-based agent from being misled. Consequently, we constructed $r_C$ heuristically to encourage closer alignment between the positive self-evaluation rate and the actual task completion rate, as shown in Equation \ref{eq:cabiliration-reward}.

Similar to the construction of the proxy policy shown in Appendix \ref{app:theory}, we can show that the optimal trajectory distributions under KL regularization for $r_E$ and $r_P$ can be defined respectively as
\begin{equation}
\begin{aligned}
    P_E(\tau) \propto \exp(r_E(\tau)), \ \ P_P(\tau) \propto \exp(r_P(\tau)).
\end{aligned}
\end{equation}
As a result, using adapted absolute differences between $p_\text{real}$ and $p_\text{self-eval}$ is able to pulling the optimization direction of $r_P$ towards the real environmental direction $r_E$.

Figure \ref{fig:heuristic} (left) illustrates the function of $r_C$ with the absolute difference between the positive self-evaluation rate and the task completion rate as the independent variable. As depicted in the figure, $r_C$ penalizes significant inconsistencies between the two rates, while not requiring them to be perfectly identical, recognizing that some degree of difference is expected. Figure \ref{fig:heuristic} (right) gives an example that how different self-evaluations and their corresponding rewards (rewards for self-evaluation) are rewarded when the actual task completion rate is 0.7.

\section{Input Representations}
\label{app:input-representation}

TextWorld environments is able to provide a rich set of environmental information at each step, including reachable items, admissible commands, current inventory, and so on. However, to ensure the generality of the challenge, we provide the LLM-based agents with only the feedback received in response to their actions, and no other environmental information is supplied as input.

Although we set a maximum of $20$ steps for the agents, this could still lead to excessive memory consumption. To mitigate this, during both the generation and adaptation, we retain only the most recent $10$ rounds of conversation. Furthermore, during adaptation, we may retain fewer rounds of history so that the prompt always contains fewer than $1280$ tokens.

\section{Implementation Details}
\label{app:implementation}

\subsection{Hardware and Framework}

All model adaptation is performed using NVIDIA A800 80GB GPUs, while inference is run on NVIDIA TITAN RTX 24GB GPUs.

All adaptation methods, including SFT-based methods (SFT, RFT), DPO-based methods (online DPO, ETO, StepAgent), and PPO-based methods (PPO, MISE), are implemented using the \texttt{trl} library\footnote{\url{https://github.com/huggingface/trl}}. For trajectory-wise DPO (ETO), we developed a custom data collator that ignore all token labels except those belonging to assistant messages, thereby focusing the learning process exclusively on the assistant's behaviors.

\begin{table*}[htb!]
    \centering
    \begin{tabular}{lccc}
        \toprule
        \textbf{Hyper-Parameters} & \textbf{SFT-Based Methods} & \textbf{PPO-Based Methods} & \textbf{DPO-Based Methods} \\
        \midrule

        LoRA rank & $64$ & $64$ & $64$ \\
        LoRA $\alpha$ & $64$ & $64$ & $64$ \\
        LoRA modules & all linear & all linear & all linear \\
        LoRA dropout & $0.0$ & $0.0$ & $0.0$ \\
        \midrule

        learning rate & $3e-6$ & $1e-5$ & $3e-7$ \\
        lr scheduler & linear & constant & linear \\
        warmup ratio & $0.1$ & $0.0$ & $0.1$ \\
        epoch & $1$ & $3$ & $1$ \\
        batch size & $256$ & $32$ & $32$ \\
        \midrule

        PPO KL coefficient & - & $0.1$* & - \\
        DPO $\beta$ & - & - & $0.3$ \\
        \bottomrule
    \end{tabular}
    \caption{Hyperparameters for the experiments. SFT-based methods include SFT, RFT; PPO-based methods include PPO, MISE; DPO-based methods include online DPO, ETO, StepAgent. The asterisk symbol denotes that, we use KL coefficient of $0.1$ with LLaMA3 and Gemma2. With Qwen2, we use higher coefficient of $0.2$. This adjustment was made because Qwen2, being initially less capable, tends to diverge at an early training stage with the lower KL coefficient. Increasing the KL coefficient helps to stabilize the training process for the model.}
    \label{tab:hyper-parameters}
\end{table*}

\subsection{Hyperparameters}

The hyperparameters used in the experiments are listed in Table \ref{tab:hyper-parameters}.

\subsection{Data and Computational Cost for Label-Free Methods}

\paragraph{RFT}
We take the hypothesis on valid set ($222$ environments) of the original model, and filter out all the trajectories yielding a score of zero, resulting in a dataset containing $1,781$ prompt-action pairs. Fine-tuning on the dataset takes less than $20$ minutes on one NVIDIA A800. 

\paragraph{Online DPO}
For each of the $222$ valid environments, two trajectories are generated using the original model with a temperature setting of $1.0$. A pairwise comparison is then performed between the two trajectories within each environment. If one trajectory has a superior score, both trajectories are added into the preference dataset. This process yielded a preference dataset containing $84$ trajectories. Subsequent trajectory-wise DPO utilizing this dataset costs less than $20$ minutes.

\paragraph{PPO and MISE}
These two methods directly use the $222$ valid environments to conduct label-free reinforcement learning. Vanilla PPO takes about $28$ hours, and MISE takes about $48$ hours. For Mise on Gemma2, we additionally adopt $4$-bit quantization to fit the model in to 80GB GPU memory, and the quantized RL process takes about only $22$ hours.

\subsection{Data and Computational Cost for Label-Supervised Methods}

\paragraph{SFT}
We directly use the walkthrough labels of the training set ($4440$ trajectories) to conduct SFT, which takes about $9$ hours.

\paragraph{SFT for MISE}
In addition to walkthrough labels, GPT-4o is employed to evaluate all labels, and these evaluations are incorporated into the SFT dataset to improve the self-evaluation capability of the model. The inclusion of this supplementary data does not significantly increase the training load. The SFT process still takes about $9$ hours on one NVIDIA A800 GPU.

\paragraph{ETO}
The process of ETO is quite similar to the trajectory-wise online DPO, with three differences: (1) ETO is based on the SFT model, (2) the data is collected based on training set instead of valid set, and (3) the preferred trajectory is derived from walkthrough labels rather than a superior generated hypothesis. ETO costs about $4$ hours.

\paragraph{StepAgent}
StepAgent is also based on SFT model. To collect StepAgent dataset, we firstly operate a teacher-forcing generation on training set, which is to collect the data of how the model acts differently to the reference label given the same history. The teacher-forcing generation on training set takes about $11$ hours on one NVIDIA TITAN RTX. Then, the collected step-wise preference data is used for step-wise DPO, which costs about $32$ hours.

\section{ScienceWorld and WebShop Settings}
\label{app:more-experiments}

In our research, the primary experiments were conducted on FTWP \cite{FirstTextWorldProblems2019} to investigate core hypotheses. Meanwhile, we conduct additional experiments to assess the generalizability of our findings across diverse agent tasks. Specifically, we adopt two datasets: ScienceWorld \cite{wang2022scienceworld} and WebShop\cite{yao2022webshop}. The ScienceWorld dataset simulates an educational platform, encapsulates structured interactions between learners and scientific content, including problem-solving trajectories, conceptual queries, and engagement patterns across various STEM domains. On the other hand, the WebShop dataset simulates an e-commerce platforms, requiring an agent to navigate and purchase according to given demands and budgets. By incorporating these datasets, we aim to demonstrate that MISE transcends domain-specific constraints, effectively addressing agent tasks with different features.

On the two additional tasks, all the preprocessing, training, and decoding details are the same to those with FTWP, except that (1) the maximum step on WebShop task is set to $10$ instead of $20$, as the environment feedback from WebShop environment is significantly longer than the others, (2) The KL coefficient of Qwen2 PPO training is $0.1$ instead of $0.2$, and (3) the environmental rewards in WebShop are "w/o IR" defined in Equation \ref{eq:ir}, as the intermediate rewards are not given in this environment. Moreover, regarding the evaluation of ScienceWorld and WebShop, all task instances within have the same maximum score, making the $\text{MicroRate}$ and $\text{MacroRate}$ identical, so we only report $\text{MicroRate}$ here.

\section{Insights into Improvement over Supervision}
\label{app:insights}

Experiments in Table \ref{tab:supervised-exp} shows that even after supervised training on a large dataset, LLM-based agents can still benefit from label-free MISE reinforcement learning on a relatively small dataset. This is reasonable, as the introduction of MISE also introduces new data. Meanwhile, we offer a specific perspective on bias in training labels to explain this improvement.

Unlike some tasks with one single correct solution, text-based games often allow for multiple winning trajectories. However, even among these correct trajectories, some may be considered suboptimal due to the inclusion of inefficient or redundant actions. A common flaw observed in our training labels is the consistent attempt to try to open a door when moving between rooms connected by a door, regardless of whether the door is already open.

When applying SFT with these suboptimal labels, the fine-tuned model will exhibit flaw in its generated trajectories as well. In test set, the SFT-trained agent attempts to open already open doors for $909$ times, a stark contrast to the original LLaMA3-8B model, which only does that for $15$ times.

However, when MISE is applied to the SFT-trained agent, the number of such attempts drops to 665. This reduction may occur because the self-evaluation mechanism can identify the error based on the environmental feedback indicating that the door is already open, and subsequently provide negative feedback for such actions, thus discouraging them.

\end{document}